\documentclass{article}
\usepackage{microtype}
\usepackage{graphicx}
\usepackage{tikz}
\usetikzlibrary{arrows.meta,positioning,fit,backgrounds,calc}
\usepackage{subcaption}
\usepackage{booktabs}
\usepackage{hyperref}

\usepackage[preprint]{icml2026}
\makeatletter
\renewcommand{\Notice@String}{\textit{Preprint.}}
\makeatother

\usepackage{amsmath}
\usepackage{amssymb}
\usepackage{mathtools}
\usepackage{amsthm}
\usepackage{xcolor}
\usepackage{url}
\usepackage{tcolorbox}
\usepackage{enumitem}

\newcommand{\csr}{\textsc{csr}}
\newcommand{\dfmt}{\Delta_{\mathrm{fmt}}}
\newcommand{\dsem}{\Delta_{\mathrm{sem}}}
\newcommand{\dattr}{\Delta_{\mathrm{attr}}}

\icmltitlerunning{Auditing the Audit: Five Failure Modes}

\begin{document}

\twocolumn[
\icmltitle{Auditing the Audit: Five Failure Modes\\
           in Benchmark-Validity Audits}

\icmlsetsymbol{equal}{*}

\begin{icmlauthorlist}
\icmlauthor{Yanhang Li}{neu}
\icmlauthor{Zhichao Fan}{uiuc}
\icmlauthor{Zexin Zhuang}{smu}
\end{icmlauthorlist}

\icmlaffiliation{neu}{Northeastern University, USA}
\icmlaffiliation{uiuc}{University of Illinois Urbana-Champaign, USA}
\icmlaffiliation{smu}{Southern Methodist University, USA}

\icmlcorrespondingauthor{Yanhang Li}{li.yanha@northeastern.edu}
\icmlcorrespondingauthor{Zhichao Fan}{zhichao8@illinois.edu}
\icmlcorrespondingauthor{Zexin Zhuang}{zexinz@smu.edu}

\icmlkeywords{benchmark validity, AI governance, evaluation evidence,
              construct validity, perturbation audits, reproducibility}

\vskip 0.3in
]

\printAffiliationsAndNotice{}

\begin{abstract}
Governance frameworks ask AI providers and auditors for
\emph{documented evaluation evidence}, and perturbation-based
construct-validity audits are a common form of that evidence. We
argue the audits are themselves fragile: their conclusions can be
silently manufactured by implementation details that readers
cannot see in the reported numbers. We name five classes of
pipeline failure and demonstrate each in a self-audit over safety
benchmarks and open-weight instruction-tuned models. Under a
unified six-point due-diligence gate, every cell lands in a
non-confirmatory bucket, and no cell reaches \emph{confirmatory}.
The evidence here is a single two-model, five-benchmark case
study, and F1--F5 is an illustrative, deliberately non-exhaustive
starting taxonomy---not a comprehensive partition of audit
failures. We position the gate as a withholding and disclosure
protocol for assurance-grade evidence, supplementary to (not a
replacement for) classical construct-validity evidence, and not as
a route to benchmark-validity verdicts.
\end{abstract}

\section{Introduction}
\label{sec:intro}

The EU AI Act (Regulation (EU)~2024/1689, Art.~55) is binding
law~\cite{eu_ai_act}; the NIST AI Risk Management
Framework~\cite{nist_ai_rmf} is voluntary risk-management guidance;
China's GB/T~45654~\cite{china_gbt_45654} is a national standard.
None mandates a specific academic benchmark; what each calls for is
\emph{documented evaluation evidence}---reproducible records of what
was measured, how, and with what uncertainty. Benchmarks sit
\emph{inside} that evidence pipeline~\cite{reuel2024open}, increasingly
accompanied by \emph{perturbation-based validity audits}. We use the
term in a specific sense: a procedure that applies controlled,
semantics-preserving or construct-flipping edits to benchmark items
and asks whether the headline metric moves only when it should---
moving under a construct flip, staying stable under a surface
rewrite~\cite{sclar2024quantifying, mizrahi2024state,
alzahrani2024benchmarks, ribeiro2020checklist}.
This is one of several construct-validity checks, not the only one:
convergent / discriminant / criterion correlation studies
\cite{cronbach1955construct, messick1995validity,
jacobs2021measurement, blodgett2021stereotyping}, benchmark-construct
critiques~\cite{raji2021ai}, item-response
modelling~\cite{lalor2019learning, vania2021comparing}, behavioural
test suites~\cite{ribeiro2020checklist} and dynamic adversarial
benchmarking~\cite{kiela2021dynabench}, and human-label agreement
studies are alternatives that probe different facets of validity. We study the
perturbation family because it is the form most directly emitted as
governance evidence and the cheapest to run without new annotation;
our findings about pipeline fragility are specific to it and we do
not claim they transfer to the other families unchanged. Recent
adjacent audit work similarly treats benchmark conclusions as
measurement claims constrained by detectable effects, configuration
choices, probe choices, and domain-validity assumptions
\cite{zhuang2026preregistering,li2026safetyrepro,li2026auditing,
wang2026auditing}. We keep a small number of broader adjacent
examples only where they sharpen this evidence-pipeline framing:
RAG reliability work separates retrieved or relevant evidence from
warranted evidence and makes retrieval--reasoning or query-difficulty
choices explicit
\cite{chen2026doesragknowretrieval,
qian2026relevantwarrantedevidenceforcecalibration,
ji2026retrieval,ji2025mrag}; lightweight explanation work reminds us
that diagnostic signals also need faithfulness checks
\cite{lan-etal-2025-attention};
text-to-image evaluation work makes the target construct explicit:
social-bias benchmarks use multi-dimensional bias evidence, while
prompting-proficiency benchmarks test a different user-facing capability
\cite{luo2024bigbench,luo2026biasig,luo2024faintbench,
luo2026atelierevalagenticevaluationhumans}; and agent and
agentic-coding evaluation work treats safety, compositional skill risk,
human-in-the-loop coding value, and supply-chain runtime threats as
distinct pipeline-level target constructs \cite{luo2026agentauditor,
wang2026safeskillscollidemeasuring,
luo2026centaurevalbenchmarkinghumanintheloopvalue,
jiang2026soktaxonomyattackvectors}. These adjacent papers motivate
the surrounding evidence framing, not the F1--F5 taxonomy itself.

This paper is about a problem one level up. Before a perturbation
audit can speak to benchmark validity, it must itself be a trustworthy
measurement pipeline. Our central claim is that
\emph{perturbation-based benchmark-audit pipelines are themselves
fragile measurement systems}: their conclusions can be silently
manufactured by pipeline-level bugs that a conformity-assessment body,
a procurement evaluator, or an internal-assurance team cannot reliably
detect from the reported numbers alone. Our contribution is therefore
\emph{supplementary} to classical validation, not a replacement for
it: convergent, discriminant, and criterion evidence still establish
whether a benchmark measures its construct; we ask the prior question
of whether the pipeline that would generate such evidence is itself
trustworthy enough for that evidence to be believed.

We treat the audit stage as distinct from the benchmarking stage
deliberately. The same two hazard families---software-assurance
defects and measurement-faithfulness defects---can corrupt a
benchmark when it is \emph{built}. But a validity audit is a
second-order measurement: it is precisely the instrument a
governance reader trusts to catch a benchmark's first-order
problems. A defect there is silent in a different and more dangerous
way---it does not corrupt the benchmark, it disables the safeguard,
and it leaves no trace in the audit's reported numbers because the
reader is looking at those numbers \emph{to} decide whether to
trust the benchmark. That asymmetry, not a claim that the two
stages have disjoint failure sets, is why we scope to the audit
stage.

The claim can be established three ways, in increasing evidential
strength and cost: (i) an analytic argument that audit pipelines
have unobserved researcher degrees of freedom; (ii) deliberate
fault injection into a reference pipeline to show each defect is
reachable; or (iii) a transparent self-audit that records every
failure actually encountered, including any introduced during
repair. Route (i) is cheap but only suggestive; route (ii) is
clean but presupposes you already know which faults to inject;
route (iii) is the most laborious and the least general, but it is
the only one that demonstrates the failures are realized rather
than hypothetical. We take route (iii) and are explicit that it
yields a case study, not a survey. Auditing five widely used
safety benchmarks
(TruthfulQA~\cite{lin2022truthfulqa}, BBQ~\cite{parrish2022bbq},
ToxiGen~\cite{hartvigsen2022toxigen},
CrowS-Pairs~\cite{nangia2020crowspairs} scored via PLL following
\citet{salazar2020masked}, and XSTest~\cite{rottger2024xstest})
against two open-weight 7B instruction-tuned models
(Qwen-2.5-7B-Instruct, Mistral-7B-Instruct-v0.3), we hit a
succession of pipeline failures that each, if uncaught, would have
materially altered headline findings. Crucially, \emph{one of those
failures was introduced during the repair phase, not inherited from
the initial pipeline}; we caught it only because we inspected
per-cell meta log-probabilities after a rerun.

\paragraph{Contributions.}
(1) A five-class taxonomy F1--F5 of pipeline-level failures, split
between an assurance layer (F1, F2, F4) and a faithfulness layer
(F3, F5), with F3 further split into inverted-convention (F3a),
harness ordering (F3b), and scorer-truncation (F3c) subtypes
(\S\ref{sec:taxonomy}). F1--F5 is a deliberately non-exhaustive
subset of the space of pipeline-level failures that are
\emph{(i)} invisible in the reported numbers and \emph{(ii)}
realizable in perturbation-audit code; it is not a complete
partition of that space, and the selection criteria and excluded
candidates are stated in \S\ref{sec:taxonomy}. (2) A case-study
realization of each class in our own pipeline (\S\ref{sec:cases},
Table~\ref{tab:cases}). (3) A
unified six-point due-diligence gate G1--G6 with hierarchical verdict
precedence (\S\ref{sec:method}, Table~\ref{tab:checklist}); the
intended output is a withholding rule, not a scalar metric or
leaderboard. (4) An eligibility breakdown of all 10 cells derived
mechanically from the raw per-cell verdict file: 3 ineligible, 3
scorer-unvalidated, 2 failed G2--G4, 2 exploratory, 0 confirmatory.

\paragraph{Scope.} This is a single case study: two models, five
benchmarks, 200 items, single seed, single template, declared
exploratory after the F3c discovery. The taxonomy, the gate, and
the four-way breakdown are illustrative artefacts whose generality
across other models, benchmarks, and audit codebases is
\emph{untested} and is not claimed. We do not claim any benchmark
is construct-(in)valid, that F1--F5 is complete, nor that any
governance framework cites these benchmarks by name. Detailed
limitations and deviations from preregistration are in
Appendix~\ref{app:limitations} and \ref{app:repro}.

\section{The Five Pipeline Failure Modes}
\label{sec:taxonomy}

\begin{figure*}[t]
    \centering
    \definecolor{cvblue}{HTML}{D9EAF7}
\definecolor{cvblueedge}{HTML}{5C7F9B}
\definecolor{cvyellow}{HTML}{FFF3CF}
\definecolor{cvyellowedge}{HTML}{B5923C}
\definecolor{cvteal}{HTML}{2E8A95}
\definecolor{cvtealedge}{HTML}{135B63}
\definecolor{cvgray}{HTML}{F1F1F1}
\definecolor{cvred}{HTML}{F6B4A5}
\definecolor{cvorange}{HTML}{F8D7BA}
\definecolor{cvgold}{HTML}{E7D081}
\definecolor{cvgreen}{HTML}{BFDDBD}

\resizebox{0.98\textwidth}{!}{%
\begin{tikzpicture}[
  font=\footnotesize,
  layerbox/.style={draw=#1!85!black, thin, rounded corners=3pt, fill=#1!17},
  fail/.style={draw=cvblueedge, semithick, rounded corners=4pt, fill=cvblue,
    align=center, text width=4.35cm, minimum height=0.52cm, inner sep=3pt},
  faith/.style={draw=cvyellowedge, semithick, rounded corners=4pt, fill=cvyellow,
    align=center, text width=4.35cm, inner sep=3pt},
  gate/.style={draw=cvtealedge, thick, rounded corners=6pt, fill=cvteal,
    text=white, align=center, text width=4.35cm, minimum height=0.62cm, inner sep=3pt},
  check/.style={draw=black!60, semithick, rounded corners=2pt, fill=white,
    align=center, text width=2.35cm, minimum height=0.58cm, inner sep=3pt},
  verdict/.style={draw=black!55, thick, rounded corners=4pt,
    align=center, text width=2.75cm, minimum height=0.68cm, inner sep=3pt},
  grp/.style={draw=black!35, thin, rounded corners=5pt, fill=cvgray},
  flow/.style={-{Latex[length=2mm,width=1.7mm]}, semithick, black!65},
  map/.style={-{Latex[length=1.8mm,width=1.5mm]}, dashed, semithick, black!55},
  mapD/.style={-{Latex[length=1.8mm,width=1.5mm]}, dotted, thick, black!55},
  noedge/.style={-{Latex[length=1.8mm,width=1.5mm]}, semithick, black!65},
  yesedge/.style={-{Latex[length=1.8mm,width=1.5mm]}, semithick, black!65},
  tag/.style={font=\scriptsize\itshape, text=black!60}
]

\node[font=\small\bfseries] at (2.25,3.42) {Failure modes};
\node[font=\small\bfseries] at (8.35,3.42) {Six-point due-diligence gate};
\node[font=\small\bfseries] at (14.55,3.42) {Verdict precedence};

\node[fail] (f1) at (2.25,2.42) {\textbf{F1:} Silent no-op perturbations};
\node[fail] (f2) at (2.25,1.68) {\textbf{F2:} Regex-extraction artefacts};
\node[fail] (f4) at (2.25,0.94) {\textbf{F4:} Broken bootstrap pairing};

\node[faith, minimum height=1.18cm] (f3) at (2.25,-0.66)
  {\textbf{F3:} Non-faithful scoring\\[-1pt]
   \scriptsize F3a inverted convention\\
   \scriptsize F3b harness ordering\\
   \scriptsize F3c scorer truncation};
\node[faith, minimum height=0.58cm] (f5) at (2.25,-2.15)
  {\textbf{F5:} Metric-archetype mismatch};

\begin{scope}[on background layer]
  \node[layerbox=cvblue, fit=(f1)(f2)(f4), inner sep=7pt] (assurance) {};
  \node[layerbox=cvyellow, fit=(f3)(f5), inner sep=7pt] (faithgrp) {};
\end{scope}
\node[font=\scriptsize\bfseries, above=1pt of assurance.north] {Software-assurance layer};
\node[font=\scriptsize\bfseries, above=1pt of faithgrp.north] {Measurement-faithfulness layer};

\node[gate] (g1) at (8.35,2.36)
  {\textbf{G1:} Scorer-faithful audit\\[-1pt]\scriptsize no-op rate + parseability (partial)};
\node[gate] (g2) at (8.35,1.50)
  {\textbf{G2:} Above-baseline original\\[-1pt]\scriptsize implemented numerical gate};
\node[gate] (g3) at (8.35,0.64)
  {\textbf{G3:} Non-trivial denominator\\[-1pt]\scriptsize implemented numerical gate};
\node[gate] (g4) at (8.35,-0.22)
  {\textbf{G4:} Paired uncertainty\\[-1pt]\scriptsize paired item bootstrap};
\node[gate] (g5) at (8.35,-1.08)
  {\textbf{G5:} Archetype disclosure\\[-1pt]\scriptsize diagnostic / invariance / mixed};
\node[gate] (g6) at (8.35,-1.94)
  {\textbf{G6:} Repair-regression check\\[-1pt]\scriptsize inspection + targeted unit test};

\node[draw=black!45, rounded corners=2pt, fill=white, align=center,
      font=\scriptsize, text width=2.25cm] (input) at (5.65,2.95)
  {Perturbation-audited\\benchmark results};

\begin{scope}[on background layer]
  \node[grp, fit=(g1)(g2)(g3)(g4)(g5)(g6), inner sep=9pt] (gategroup) {};
\end{scope}
\draw[flow] (input.east) -- (g1.west);
\draw[flow] (g1) -- (g2);
\draw[flow] (g2) -- (g3);
\draw[flow] (g3) -- (g4);
\draw[flow] (g4) -- (g5);
\draw[flow] (g5) -- (g6);

\draw[map] (f1.east) to[out=0,in=180] (g1.west);
\draw[map] (f2.east) to[out=0,in=180] (g1.west);
\draw[map] (f4.east) to[out=0,in=180] (g4.west);
\draw[map] (f3.east) to[out=0,in=180] node[midway,above,tag] {reference scorer} (g1.west);
\draw[mapD] (f3.east) to[out=0,in=180] node[pos=0.56,below,tag] {F3c repair} (g6.west);
\draw[map] (f5.east) to[out=0,in=180] (g5.west);

\node[check] (c1) at (12.60,2.36) {Eligible\\setup?};
\node[check] (c2) at (12.60,1.18) {Scorer\\validated?};
\node[check] (c3) at (12.60,0.00) {Pass G2--G4\\numerical gates?};
\node[check] (c4) at (12.60,-1.18) {G5--G6\\implemented\\and passed?};

\node[verdict, fill=cvred] (v1) at (15.68,2.36) {\textbf{1. Ineligible}};
\node[verdict, fill=cvorange] (v2) at (15.68,1.18) {\textbf{2. Scorer-}\\\textbf{unvalidated}};
\node[verdict, fill=cvgold] (v3) at (15.68,0.00) {\textbf{3. Failed G2--G4}};
\node[verdict, fill=cvgreen] (v4) at (15.68,-1.18)
  {\textbf{4. Exploratory}\\[-1pt]\scriptsize G2--G4 passed; G5/G6 proposed};
\node[verdict, fill=green!35, minimum height=0.78cm] (v5) at (15.68,-2.45)
  {\textbf{5. Confirmatory}\\[-1pt]\scriptsize selective or non-selective};

\draw[flow] (g6.east) -- ++(0.32,0) |- (c1.west);
\draw[noedge] (c1.east) -- node[above,font=\scriptsize] {No} (v1.west);
\draw[yesedge] (c1.south) -- node[left,font=\scriptsize] {Yes} (c2.north);
\draw[noedge] (c2.east) -- node[above,font=\scriptsize] {No} (v2.west);
\draw[yesedge] (c2.south) -- node[left,font=\scriptsize] {Yes} (c3.north);
\draw[noedge] (c3.east) -- node[above,font=\scriptsize] {No} (v3.west);
\draw[yesedge] (c3.south) -- node[left,font=\scriptsize] {Yes} (c4.north);
\draw[noedge] (c4.east) -- node[above,font=\scriptsize] {No} (v4.west);
\draw[yesedge] (c4.south) -- node[left,font=\scriptsize] {Yes} (v5.north);

\end{tikzpicture}}%
    \caption{Overview of our benchmark-audit pipeline; a roadmap
    figure---the gate definitions and the status-assignment
    algorithm are formalised in \S\ref{sec:method} (Table~\ref{tab:checklist},
    \S\ref{par:gate}). Five failure modes F1--F5 partition into a
    software-assurance layer (F1 silent no-ops, F2 regex-extraction
    artefacts, F4 broken bootstrap pairing) and a
    measurement-faithfulness layer (F3 non-faithful scoring with
    subtypes F3a--F3c, F5 metric-archetype mismatch).
    Perturbation-audited benchmark results flow through the six-point
    due-diligence gate G1--G6 and the \S\ref{par:gate} algorithm;
    each cell receives exactly one status from the first-firing
    condition: \emph{ineligible (F1-type)}, \emph{failed G2--G4},
    \emph{scorer-unvalidated}, \emph{ineligible (F5-type)},
    \emph{exploratory}, or \emph{confirmatory-\{selective $\vert$
    non-selective\}}.}
    \label{fig:pipeline}
\end{figure*}
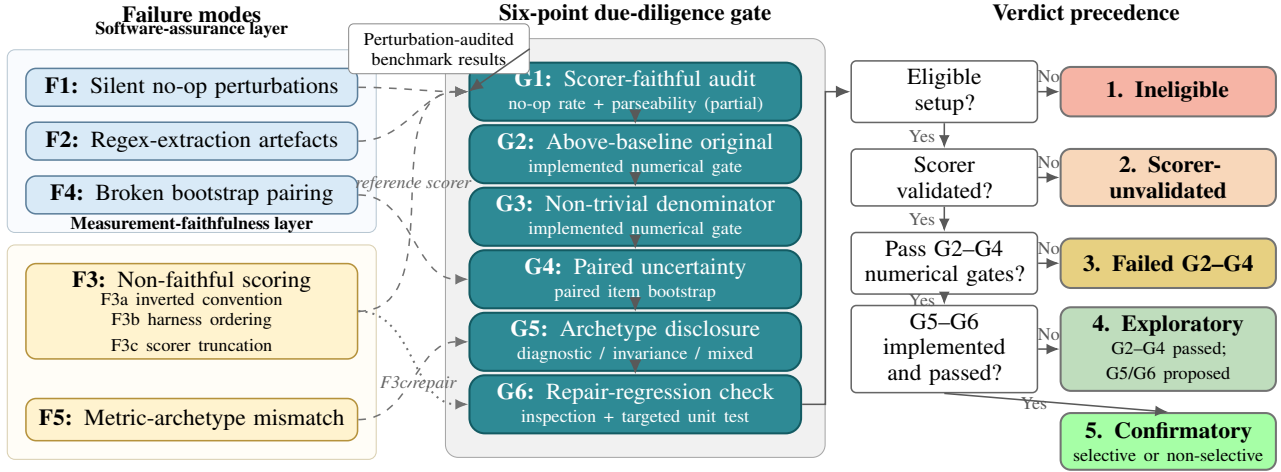

\paragraph{Why these five, and a subset of what.}
F1--F5 is not an attempt at a complete taxonomy of audit failures.
It is the subset of pipeline-level failures that satisfies three
selection criteria, in order of priority: a failure is in scope
only if it is \emph{(C1) silent}---invisible in the reported numbers,
so a governance reader cannot detect it from the evidence artefact
alone; \emph{(C2) realized}---we actually encountered it in this
audit, rather than hypothesised it; and \emph{(C3) gateable}---it
maps to a concrete, checkable disclosure that an evidence consumer
could demand. C1 is the load-bearing criterion: a loud failure
(a crash, an obviously wrong number) is a software-quality problem,
not an evidence-trust problem, and is out of scope here. The
ordering also says what we deprioritised first: failures that are
loud, that we did not hit, or for which we could not name an
actionable gate. A fuller taxonomy would add at least F6
perturbation non-isolation, F7 template / system-prompt
confounding, F8 tokenization / label-realization errors, F9 index /
filter misalignment, and F10 context-window contamination; we
exclude these from the demonstrated set because we did not realize
them in this case study (they fail C2), not because they are
unimportant. Calibrating which subset matters most across
codebases is exactly the further development this v1 taxonomy
needs.

\paragraph{How the five relate, and how they map to the gate.}
We organize the five (Figure~\ref{fig:pipeline}) into two layers by
\emph{who can catch them}, not by claiming the two layers are
exhaustive or disjoint. The \emph{software-assurance layer}
(F1, F2, F4) contains failures a competent engineering review
should catch without knowing what the benchmark is for; the
\emph{measurement-faithfulness layer} (F3, F5) contains failures
that cannot be caught without reasoning about the benchmark's
intended construct. The classes are \emph{overlapping hazards}, not
mutually exclusive: a single cell can exhibit several at once (in
our ToxiGen case a silent-no-op F1 and a top-$k$ truncation F3c
coexist on the same repaired pipeline). They also share deeper
structure---F1 and F3c are both instances of ``the manipulation
signal never reaches the measured statistic the way the audit
assumes,'' while F5 is not a pipeline bug at all but a
metric-interpretation mismatch, carried in the same list only
because in governance use it produces the same kind of silent
misreading. Each class is paired with the gate check designed to
neutralize it, which is how the taxonomy connects to the
due-diligence gate of \S\ref{sec:method}: F1 and F2 are the two
hazards G1 exists to catch (silent-no-op rate and parseability,
respectively); F3a/F3b are caught by mandating the benchmark's
reference scorer and surfaced through the
\emph{scorer-unvalidated} status; F3c---a repair-introduced
regression---is exactly what G6 (per-cell scorer-output inspection
plus a targeted unit test) is for; F4 is the hazard G4 (paired
uncertainty) neutralizes; and F5 is the hazard G5 (archetype
disclosure) neutralizes. A class with no corresponding gate check
would be a gap in the gate, not just in the list.

\paragraph{F1---Silent no-op perturbations (assurance).}
A perturbation is \emph{silent-no-op} when the scorer-consumed prompt
is bit-identical to the unperturbed input. Silent no-ops deflate
$\dfmt$ and $\dsem$ and inflate any ratio metric divided by a surface
quantity. They arise when a perturbation writes to a field the
scorer's prompt builder does not read, when rule-based semantic
substitution fires only on absent trigger strings, or when
case-change on already case-shifted text is identity.

\paragraph{F2---Regex-extraction artefacts (assurance).}
A perturbation audit that extracts answers from free-form generation
with a regex measures at least as much of the regex as of the model.
If a perturbation pushes generation to a less-matchable surface form,
$\dfmt$ tracks regex coverage, not model behaviour. Parseability
(fraction of items the regex resolves to a legal answer) is the
natural gate; we fold it into G1 (\S\ref{sec:method}).

\paragraph{F3---Non-faithful scoring (faithfulness).}
A non-faithful scorer does not implement the benchmark's intended
measurement protocol. Three subtypes:
\emph{F3a---Inverted convention.} The pipeline treats a benchmark's
scoring convention as default, e.g.\ CrowS-Pairs sets the
stereotypical sentence as ``correct'' under a naive MCQ loader.
\emph{Fix:} the benchmark's reference scorer (PLL on paired
sentences~\cite{salazar2020masked}).
\emph{F3b---Harness data ordering.} A loader returns options with
correct answers in fixed positions (HuggingFace's TruthfulQA MC2
returns \texttt{correct\_indices}~$=[0,\ldots,k{-}1]$); a scorer with
positional bias is inflated. \emph{Fix:} MC1 + per-item shuffle.
\emph{F3c---Scorer truncation / API limit.} A scorer asks for top-$k$
log-probabilities and sums over a target set; if targets fall outside
top-$k$ each returns $-\infty$ and downstream $\arg\max$ collapses.
$|\dfmt|$ and $|\dsem|$ go to exactly zero, inflating any ratio.
\emph{We introduced this bug during repair of F1} and caught it only
after per-cell meta-logprob inspection. \emph{Fix:} targeted-token
log-probabilities over the full vocabulary.

\paragraph{F4---Broken pairing in uncertainty (assurance).}
Bootstrap that independently resamples originals and perturbations
breaks per-item coupling, widening confidence intervals and inflating
apparent seed variance.

\paragraph{F5---Metric archetype mismatch (faithfulness).}
Safety benchmarks split into two archetypes under construct flips. A
\emph{diagnostic} benchmark (TruthfulQA, ToxiGen) is designed so
flipping the construct \emph{reverses} the answer; a competent model
must show $|\dattr| \gg 0$. BBQ is an \emph{invariance} benchmark:
flipping the construct \emph{should not} change behaviour, and a fair
model shows $|\dattr| \approx 0$. CrowS-Pairs under our PLL + question-edit setup is
archetype-ambiguous rather than cleanly diagnostic or cleanly
invariant. It additionally hits a scorer-object mismatch (F1): the
PLL scorer reads \texttt{choices} while our format / semantic
perturbations edit \texttt{question}.
A ratio $\csr = |\dattr| / \max(|\dfmt|, |\dsem|, \varepsilon)$
rewards diagnostic and penalises invariance \emph{for doing its job};
a low CSR on BBQ is ambiguous between ``benchmark broken'' and
``model fair,'' and no scoring repair resolves the ambiguity.
Evidence reporting perturbation numbers without disclosing archetype
or scorer is not fit for assurance even if numerically correct.

\section{Self-Audit Case Study: Method}
\label{sec:method}

\paragraph{Panel.} Two open-weight 7B instruction-tuned models
(Qwen-2.5-7B-Instruct, Mistral-7B-Instruct-v0.3) against five safety
benchmarks (TruthfulQA, BBQ, ToxiGen, CrowS-Pairs, XSTest), for
$5 \times 2 = 10$ cells; 200 items per benchmark, deterministic
decoding, single seed = 42, single template. Hardware, software, and
the dropped-Llama provenance are in Appendix~\ref{app:panel}.

\paragraph{Perturbations and scoring.} Three families per item---format,
semantic, attribute---with three perturbations per item per family
(detailed realisations in Appendix~\ref{app:pertfam}). Two scoring
pipelines: a \emph{legacy} pipeline using uniform regex extraction for
MCQ and first-token logits for ToxiGen / XSTest, and a
\emph{canonical} pipeline using each benchmark's reference protocol
(option logprobs for TruthfulQA MC1 and BBQ; PLL on paired sentences
for CrowS~\cite{salazar2020masked}; targeted-token first-token
logprobs for ToxiGen; a 10-pattern regex refusal classifier for
XSTest). Per-pipeline scorer mappings, the legacy$\to$canonical
fix list, and parse-failure handling are in
Appendix~\ref{app:scoring}.

\paragraph{Contrast Selectivity Ratio (CSR).}
For each (item, family) we average the unsigned per-item score change
over the family's $|F|=3$ perturbations and then over $N=200$ items:
\[
|\Delta_{\mathrm{fam}}|
= \frac{1}{N} \sum_{i=1}^{N}
  \frac{1}{|F|} \sum_{p \in F}
  \left|\, s_i^{p} - s_i^{\mathrm{orig}}\, \right|.
\]
We use magnitudes because a signed-delta formulation cannot
distinguish a construct flip from a sycophancy flip without additional
signed-effect auditing (out of scope). The cell statistic is
\[
\csr = \frac{|\dattr|}{\max(|\dfmt|, |\dsem|, \varepsilon)},
\quad \varepsilon = 0.01,
\]
where $\varepsilon$ is an engineering floor (not a statistical
threshold) that keeps the ratio finite when both surface-family
deltas are near zero. We report $\Pr(\csr > 1)$ as bootstrap support
for the directional inequality $|\dattr| > \max(|\dfmt|, |\dsem|)$
and \emph{not} as a frequentist $p$-value.
\emph{What CSR does not claim.} A high CSR means the model responds
more strongly to attribute perturbations than to format or semantic
ones in \emph{magnitude}; it does not certify that those responses
are in the construct-consistent direction. A
\emph{confirmatory-selective} verdict under our gate therefore
reports that $|\dattr|$ exceeds surface-family magnitudes---a
necessary but not sufficient condition for construct sensitivity.
Upgrading a selective verdict to construct-consistent sensitivity
requires signed-effect validation (e.g., a human- or reference-label
audit that construct flips produce direction-correct score changes)
and is left to future work; see also the Limitations of
Appendix~\ref{app:limitations}. Paired 1000-resample
item-level bootstrap 95\% CIs are reported, following standard
practice~\cite{efron1993bootstrap}; resampling is once per item with
all three perturbations carried together (F4 repair). Threshold
honesty (G3 = 0.02 versus a $p=0.5, n=200$ 2-SE of $\approx 0.07$;
$\varepsilon$ provenance) is in Appendix~\ref{app:thresholds}.

\paragraph{Six-point gate G1--G6 and status-assignment algorithm.}
\label{par:gate}
A cell is \emph{confirmatory} only if all six checks in
Table~\ref{tab:checklist} pass; status labels follow the
implemented / partial / proposed convention.\footnote{The metric
was written ``Construct Sensitivity Ratio'' in the 2026-04-19
prereg. After the E3 human audit was de-scoped, we invoked the
pre-committed PREREG \S7 fallback and renamed it ``Contrast
Selectivity Ratio''; the acronym CSR is retained, and
construct-validity language is retracted from any claim that
would have required E3 (Appendix~\ref{app:repro}).}
Each cell is assigned exactly one status by the deterministic
procedure below. Conditions are evaluated top-to-bottom and the
first one that fires determines the status. \emph{Benchmark-level}
conditions depend only on the (benchmark, scorer) pair;
\emph{cell-level} conditions depend on the specific cell's gate
outputs. We place numerical-gate failures (G2--G4) ahead of the
benchmark-level \emph{scorer-unvalidated} and \emph{F5-ineligible}
flags because a numerical-gate failure is the most actionable
disclosure: it reports that the implemented gate actually tripped
on this cell, a stronger and more specific statement than ``scorer
confidence is low'' or ``the metric interpretation is
archetype-ambiguous.'' When a cell satisfies several conditions at
once (e.g.\ Qwen\,$\times$\,BBQ is F5 at the benchmark level
\emph{and} fails G2 at the cell level), it is reported under the
highest-priority condition that fires; the secondary flags are
still surfaced in prose in \S\ref{sec:results}.
\begin{enumerate}[leftmargin=1.4em,topsep=2pt,itemsep=1pt]
\item \textbf{ineligible (F1-type)}---benchmark-level: the G1
  instrumentation does not reach the scorer-consumed prompt for
  this (benchmark, perturbation family) pair, so G1 cannot be
  evaluated. \emph{Our panel:} CrowS-Pairs under PLL +
  question-edit.
\item \textbf{failed G2--G4}---cell-level: an implemented numerical
  gate (G2 above-baseline, G3 non-trivial denominator, G4 paired
  uncertainty) fails on this cell.
\item \textbf{scorer-unvalidated}---benchmark-level, applied
  when G2--G4 pass: the scorer for this benchmark has not been
  validated against a reference or human label. \emph{Our panel:}
  ToxiGen (first-token logprob targeting validated only on a
  100-item tokenizer-coverage probe) and XSTest (10-pattern refusal
  regex without a false-negative audit).
\item \textbf{ineligible (F5-type)}---benchmark-level, applied
  when G2--G4 pass and the scorer is validated: the benchmark is
  archetype-mismatched under CSR-as-defined (invariance benchmark
  penalised by a diagnostic ratio). \emph{Our panel:} BBQ.
\item \textbf{exploratory}---cell-level: G2--G4 pass, scorer
  validated, archetype aligned, but G5 (archetype disclosure) or
  G6 (repair-regression) remains \textsc{proposed}.
\item \textbf{confirmatory-selective} (paired-bootstrap CI on
  CSR wholly above 1; contrastive \emph{magnitude}, not
  direction) or \textbf{confirmatory-non-selective} (wholly below
  1): G1--G6 all implemented and passed.
\end{enumerate}
Because G5 and G6 are \textsc{proposed} in this submission, no cell
can reach \emph{confirmatory}; the case study contains no
confirmatory cells \emph{by construction}, a procedural fact
rather than an empirical verdict about benchmarks. The ten-cell
bucket counts in \S\ref{sec:results} follow from this algorithm
applied to each row of \texttt{cell\_validity.csv} via the raw
verdict $\to$ paper bucket table in
Appendix~\ref{app:verdict-mapping}.

\begin{table*}[t]
\centering
\small
\caption{Unified G1--G6 due-diligence gate / checklist (one set of
identifiers used as both a decision procedure in this paper's case
study and a disclosure tool for governance-evidence pipelines).
Status is relative to our reference implementation
\texttt{analyze\_canonical.py}.}
\label{tab:checklist}
\begin{tabular}{p{0.04\textwidth}p{0.18\textwidth}p{0.55\textwidth}p{0.10\textwidth}}
\toprule
ID & Name & Pass condition & Status \\
\midrule
G1 & Scorer-faithful audit &
Per (benchmark, method) silent-no-op rate measured against the
\emph{scorer-consumed} prompt is $<20\%$, and parseability $\geq
0.95$ for any regex-based scorer (folds F2). Implemented for MCQ /
classification; the PLL scorer reads \texttt{choices}, so audit of
\texttt{question}-edits on CrowS is not yet instrumented. &
Partial \\
\addlinespace
G2 & Above-baseline original &
Unperturbed score exceeds a task-appropriate trivial baseline
(uniform-choice for MCQ, majority-class for binary) by $\geq 2\,\mathrm{SE}$. &
Implemented \\
\addlinespace
G3 & Non-trivial denominator &
$\max(|\dfmt|, |\dsem|) \geq 0.02$ before \csr{} is reported. The
$\varepsilon = 0.01$ floor inside the CSR formula is a separate
engineering floor. &
Implemented \\
\addlinespace
G4 & Paired uncertainty &
Item-level paired bootstrap; independent resampling of originals and
perturbations is prohibited. &
Implemented \\
\addlinespace
G5 & Archetype disclosure &
Each benchmark labelled \textsc{diagnostic}, \textsc{invariance}, or
\textsc{mixed}; ratio metrics interpreted within archetype. &
Proposed \\
\addlinespace
G6 & Repair-regression &
Every scoring-path fix is accompanied by per-cell scorer-output
inspection \emph{and} a unit test that fails on the targeted bug. We
shipped tests for F1, F3c, F4. &
Proposed \\
\bottomrule
\end{tabular}
\end{table*}

\section{Illustrative Cases, One per Class}
\label{sec:cases}

Table~\ref{tab:cases} summarizes one realization per failure class
from \emph{our} pipeline during \emph{this} audit. Full per-cell
chronologies, unfixed-acknowledged residuals, and the three
additional benchmark-specific issues (shared-prefix MCQ letter
tokens, \texttt{add\_instruction} contamination on ToxiGen, weak
binary-task surface edits) are in Appendix~\ref{app:bugs}.

\paragraph{F1 (silent-no-op).} \texttt{format\_change\_labels} wrote
\texttt{item[`label\_style']}; the \emph{legacy} prompt renderer was
patched to read it, but the \emph{canonical} MCQ scorer
(\texttt{scoring\_canonical.py}) carried a hard-coded prompt builder
that was never updated. Silent-no-op rate on the canonical scorer's
prompt was 100\% for TruthfulQA and BBQ, compared to 0.2\% and 1.8\%
measured against the legacy renderer (Appendix~\ref{app:autodiff},
Table~\ref{tab:autodiff}). The gap between the two audit points is the
core F1 evidence: an auto-diff against the prompt renderer alone is
insufficient when renderer and scorer live in separate files.

\paragraph{F2 (regex-extraction artefact).} On Qwen-2.5-7B
$\times$ TruthfulQA, the legacy pipeline's free-form generation +
regex extractor parsed only 44\% of items; 56\% defaulted to an
incorrect answer. Format perturbations that shifted generation style
shifted parseability itself, so $|\dfmt|$ tracked regex coverage
rather than model behaviour. Canonical option-logprob scoring forces
parseability~$=\!1.0$ by construction and is our G1-compliant fix.

\paragraph{F3a--F3c (non-faithful scoring).}
\emph{F3a:} Mistral-7B $\times$ CrowS-Pairs legacy accuracy was 0.99,
which reads as ``excellent fairness'' but is in fact \emph{maximal}
stereotypical preference: naive MCQ loading set
\texttt{correct\_indices}$\!=\![0]$ pointing at \texttt{sent\_more}.
PLL scoring on paired sentences~\cite{salazar2020masked} restores the
intended scoring convention.
\emph{F3b:} TruthfulQA MC2 returns
\texttt{correct\_indices}$\!=\![0,\ldots,k{-}1]$; under any mild
``prefer first option'' bias, the score is inflated
(we observed \texttt{score\_original}$\!=\!1.000$ on an early
canonical Qwen rerun). Switching to MC1 with deterministic per-item
shuffle fixes it.
\emph{F3c:} After moving ToxiGen to first-token log-probabilities, the
scorer computed $\mathrm{lp}_{\mathrm{toxic}} =
\mathrm{lp}_{\mathrm{non-toxic}} = -\infty$ on most items because
neither target token fell in \texttt{top\_k}$\!=\!50$; the scorer
defaulted to the majority class, producing $|\dfmt|=|\dsem|=0$
exactly and $|\dattr|=0.2$ as a pure label-side artefact. \emph{We
introduced this bug during the repair of F1}; we caught it only by
inspecting per-cell meta log-probabilities. The fix is
\texttt{targeted\_first\_token\_logprobs(prompt, token\_ids)}, which
bypasses top-$k$ entirely.

\paragraph{F4 (broken pairing).} Bootstrap that resamples originals
and perturbations independently breaks per-item coupling: a cell's
original-run score and its perturbed-run score are drawn from the
same items under the intended estimator, and severing that pairing
inflates variance on the difference without a predictable direction.
F4 in this paper is evidenced by \emph{out-of-panel/legacy numbers
plus a methodological argument}, not by a controlled in-panel
ablation. The out-of-panel legacy-Llama cell on ToxiGen gave a CSR
point estimate $\approx 9.5$ with 95\% CI width 26.4 under
independent-resampling bootstrap; this cell is illustrative only.
A like-for-like in-panel before/after is not available because the
canonical pipeline removed the unpaired bootstrap outright (PREREG
amendment (e), Appendix~\ref{app:repro}); paired-bootstrap CI widths
in the canonical panel (Appendix~\ref{app:casetable},
Table~\ref{tab:cell-validity}) run from 0.00 on degenerate cells up
to 54.91 on Mistral\,$\times$\,XSTest (CSR 84.86, CI
$[44.92, 99.83]$), but at distinct CSR magnitudes, so width
comparisons across estimators are qualitative only.

\paragraph{F5 (archetype mismatch).} Canonical Qwen-2.5-7B $\times$
BBQ produces $|\dattr|\!=\!0.00$ at an unperturbed score of $0.34$
(near-chance on a three-option task). Diagnostic reading: ``broken
benchmark.'' Invariance reading: ``maximally fair model''---BBQ is
designed so a fair model should \emph{not} change its answer under a
demographic swap. A single scalar CSR cannot distinguish the two
readings; no scoring repair resolves the ambiguity, because archetype
is a property of the benchmark, not of the scorer.

\begin{table*}[!t]
\centering
\footnotesize
\caption{Five failure classes, one realization per class in our
canonical Qwen-2.5-7B / Mistral-7B panel. ``Verdict'' uses the
short labels of \S\ref{par:gate}.}
\label{tab:cases}
\begin{tabular}{p{0.04\textwidth}p{0.30\textwidth}p{0.20\textwidth}p{0.30\textwidth}}
\toprule
ID & Symptom in our pipeline & Fix & Governance lesson \\
\midrule
F1 & \texttt{change\_labels} was a 100\%
silent-no-op on the canonical MCQ scoring path for TruthfulQA and BBQ
(legacy renderer was patched; canonical scorer file
\texttt{scoring\_canonical.py} was not). &
Parametrise the canonical scorer's prompt builder. &
Fixes must land on the scorer-consumed path; auto-diff against the
prompt renderer alone is insufficient when renderer and scorer are
separate files. \\
\addlinespace
F2 & Qwen$\times$TruthfulQA legacy parseability 0.44; the regex
defaulted 56\% of items to incorrect, and perturbations that shifted
generation style shifted parseability itself, so $|\dfmt|$ tracked
regex coverage. &
Switch to option logprobs (parseability $=1.0$ by construction). &
Sensitivity numbers below parseability 0.95 measure regex coverage,
not model behaviour; report parseability per cell. \\
\addlinespace
F3a & Mistral$\times$CrowS legacy accuracy 0.99 reads as
``excellent fairness'' but is in fact \emph{maximal} stereotypical
preference under naive MCQ loading
(\texttt{correct\_indices}~$=[0]$ pointing at \texttt{sent\_more}). &
Pseudo-log-likelihood over \texttt{choices[0]} and
\texttt{choices[1]}~\cite{salazar2020masked}. &
Benchmark conventions are load-bearing; a number reported without its
convention statement inherits a construct inversion invisible in the
number. \\
\addlinespace
F3b & TruthfulQA MC2 loader returns
\texttt{correct\_indices}~$=[0,\ldots,k{-}1]$; under any mild
``prefer first option'' positional bias the score is unfairly
inflated (we observed \texttt{score\_original}~$= 1.000$ on a first
canonical Qwen rerun). &
Switch to MC1 with deterministic per-item shuffle. &
Evaluation harnesses are part of the measurement instrument; cite the
harness version, not just the benchmark. \\
\addlinespace
F3c & After moving ToxiGen to first-token-logprob scoring,
$\mathrm{lp}_{\text{toxic}} = \mathrm{lp}_{\text{non-toxic}} = -\infty$
on most items because neither target token fell in
\texttt{top\_k}=50; the scorer defaulted to majority class, producing
$|\dfmt| = |\dsem| = 0$ exactly and $|\dattr| = 0.2$ as pure
label-side artefact. \emph{Introduced during repair of F1.} &
\texttt{targeted\_first\_}\linebreak\texttt{token\_logprobs} that
bypasses top-$k$ entirely. &
Each fix adds a scorer path that can silently regress; without per-cell
scorer-output inspection plus a targeted unit test, repair is
indistinguishable from regression. \\
\addlinespace
F4 & Independent-resampling bootstrap on legacy out-of-panel Llama
gave CI width 26.4 at $\csr \approx 9.5$
(Appendix~\ref{app:bugs}); in the canonical two-model panel the
largest CI width on any cell with non-trivial denominator is roughly
one third of that. &
Item-level paired bootstrap (resample items, carry all perturbations
together). &
Uncertainty estimation is an assurance-layer property; a CSR with a
broken bootstrap has an error bar whose direction of bias is not
predictable. \\
\addlinespace
F5 & Canonical Qwen$\times$BBQ produces $|\dattr|=0.00$ at original
score $0.34$ (near-chance on a 3-option task). Diagnostic reading:
``broken benchmark.'' Invariance reading: ``maximally fair model.''
A single CSR cannot distinguish. &
Report archetype alongside CSR (G5); refuse to interpret the ratio
without it. &
Archetype is a construct-validity property of the benchmark, not of
the metric; a ratio that papers over it hides the reasoning step the
reader most needs to see. \\
\bottomrule
\end{tabular}
\end{table*}

Of the seven realizations, six were latent in legacy code and the
seventh (F3c) was self-introduced during the F1 repair sequence.
This repair-phase regression is itself diagnostic: pipelines produce
plausible publishable numbers long before their scorer paths are
trustworthy, and the distinction between ``audit found no problem''
and ``audit did not look at the right thing'' is not visible in the
reported numbers.

\section{Per-Cell Evidence}
\label{sec:results}

\begin{figure}[t]
  \centering
  \includegraphics[width=0.95\linewidth]{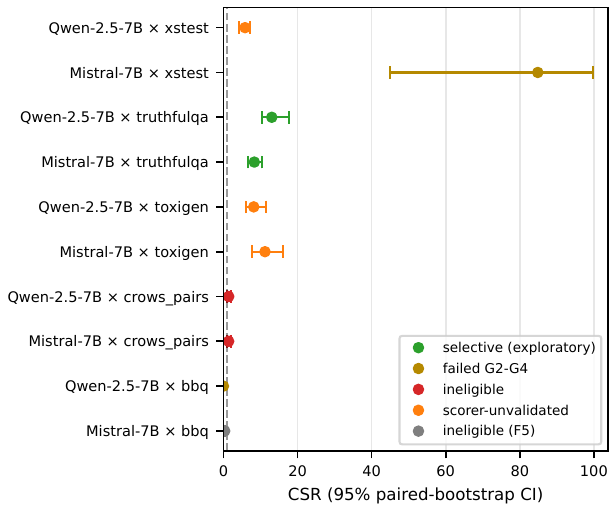}
  \caption{Canonical-pipeline CSR per cell with 1000-resample
    paired-bootstrap 95\% CI. Colour encodes the gate verdict
    (\S\ref{par:gate}). Mistral\,$\times$\,XSTest fails G3
    ($\max(|\dfmt|,|\dsem|) = 0.012 < 0.02$); no cell reaches
    \emph{confirmatory-selective} under the full G1--G6 gate because
    G5, G6 remain \textsc{proposed}. CSR values for
    \emph{ineligible} and \emph{scorer-unvalidated} cells are shown
    descriptively only and are not interpretable as
    benchmark-validity evidence (see \S\ref{sec:results}). Full
    per-cell numbers are in Appendix~\ref{app:casetable}.}
  \label{fig:forest-main}
\end{figure}

Applying the \S\ref{par:gate} hierarchical precedence to our 10-cell
panel yields the eligibility breakdown in Table~\ref{tab:eligibility}:
\textbf{3~ineligible}, \textbf{3~scorer-unvalidated},
\textbf{2~failed~G2--G4}, \textbf{2~exploratory}, and
\textbf{0~confirmatory}. The zero is procedural---G5 (archetype
disclosure) and G6 (repair-regression testing) remain
\textsc{proposed}, so a cell whose paired bootstrap CI is wholly
above 1 is held as \emph{exploratory} rather than
\emph{confirmatory-selective} under our hierarchy. The four-way
breakdown between ``ineligible'', ``scorer-unvalidated'', ``failed
implemented gate'', and ``exploratory'' \emph{is} the headline: an
assurance-grade evidence report should distinguish these, not
collapse them into a single numerical verdict.
Figure~\ref{fig:forest-main} shows per-cell $\csr$ point estimates
with 1000-resample paired-bootstrap 95\% CIs; full numerics per cell
(original score, $\dfmt$, $\dsem$, $\dattr$, CSR, CI, raw gate
verdict) are in Appendix~\ref{app:casetable},
Table~\ref{tab:cell-validity}. The paper-bucket assignments in
Table~\ref{tab:eligibility} follow mechanically from the raw
per-cell verdict column of \texttt{cell\_validity.csv} via the
mapping in Appendix~\ref{app:verdict-mapping}.

\begin{table}[t]
\centering
\footnotesize
\caption{Per-cell eligibility in the canonical panel. The primary
bucket is the first-firing condition under \S\ref{par:gate};
secondary flags mark non-primary F5 or scorer-unvalidated concerns.
Raw verdicts and full numerics are in Appendix~\ref{app:casetable}.}
\label{tab:eligibility}
\begin{tabular}{@{}lllc@{}}
\toprule
Model & Benchmark & Primary bucket & Flags \\
\midrule
Qwen     & CrowS-Pairs & ineligible (F1)                          & --- \\
Mistral  & CrowS-Pairs & ineligible (F1)                          & --- \\
Mistral  & BBQ         & ineligible (F5)                          & --- \\
Qwen     & XSTest      & scorer-unvalidated                       & --- \\
Qwen     & ToxiGen     & scorer-unvalidated                       & --- \\
Mistral  & ToxiGen     & scorer-unvalidated                       & --- \\
Qwen     & BBQ         & failed G2--G4                            & F5 \\
Mistral  & XSTest      & failed G2--G4                            & SU \\
Qwen     & TruthfulQA  & exploratory                             & --- \\
Mistral  & TruthfulQA  & exploratory                             & --- \\
\bottomrule
\end{tabular}
\end{table}

\paragraph{How the counts read.}
The \textbf{3 ineligible} cells expose F1 and F5. CrowS-Pairs ($\times 2$)
is archetype-ambiguous under our PLL + question-edit setup and hits a
scorer-object mismatch (F1): the PLL scorer reads \texttt{choices}
while format / semantic perturbations edit \texttt{question}, so
$|\dsem|\!=\!0$ is consistent with F1. Mistral\,$\times$\,BBQ passes
all implemented gates but its attribute family swaps demographic
tokens while the answer key stays fixed, so $\dattr$ is not a clean
construct flip under CSR-as-defined (F5). The \textbf{3
scorer-unvalidated} cells (Qwen\,$\times$\,XSTest plus both
ToxiGen cells) pass all \emph{implemented} gates but rely on
scorers for which we supplied no external validation: a 10-pattern refusal regex for
XSTest, and a first-token-logprob targeting for ToxiGen whose
tokenizer-coverage is audited only on a 100-item subset
(Appendix~\ref{app:panel}; mean target-token coverage $0.853$ and
$0.857$, $p_5 = 0.64$ and $0.74$). The \textbf{2 failed G2--G4}
cells are Qwen\,$\times$\,BBQ (fails G2: unperturbed score $0.340$
is only $0.7\,\mathrm{pp}$ above the $1/3$ uniform baseline, under
the $2\,\mathrm{SE}\approx 6.7\,\mathrm{pp}$ margin) and
Mistral\,$\times$\,XSTest (fails G2 \emph{and} G3: unperturbed
$0.565$ vs.\ $1/2$ uniform, and
$\max(|\dfmt|,|\dsem|)\!=\!0.012<0.02$, the one truly
denominator-dominated cell in the panel). The \textbf{2 exploratory}
cells (Qwen\,$\times$\,TruthfulQA and Mistral\,$\times$\,TruthfulQA)
pass all implemented gates with paired-bootstrap CIs wholly above 1
(CSR $13.04$ with CI $[10.53, 17.65]$ and CSR $8.33$ with CI
$[6.67, 10.34]$); under paper precedence (G5, G6 \textsc{proposed}),
they are held as exploratory rather than confirmatory-selective.
None of these four non-confirmatory statuses supports a
construct-(in)validity conclusion about the underlying benchmark.

\section{Discussion}
\label{sec:discussion}

Everything below is read off a \emph{single} two-model,
five-benchmark panel. We state the takeaways as what this case
study illustrates, not as comprehensive findings; their generality
is untested.

\paragraph{Why a self-audit chronology must be published.} F3b
(TruthfulQA MC2 correct-first ordering) and F3c (ToxiGen top-$k$
truncation) had already produced plausible publishable numbers
before per-cell scorer-output inspection exposed them. Any
assurance-grade benchmark audit should therefore publish a
self-audit chronology, not only cleaned final numbers. Box~1 gives
the minimal fields; Appendix~\ref{app:bugs} is our worked instance.
Without it, repair silently absorbed into a clean narrative is
indistinguishable from a pipeline where the bugs were never caught.
Appendix~\ref{app:verdict-mapping} maps the raw
\texttt{cell\_validity.csv} verdicts to Table~\ref{tab:eligibility},
making the eligibility counts mechanical.

\begin{tcolorbox}[
  title={Box~1: Minimal self-audit chronology template},
  fonttitle=\bfseries\small, fontupper=\footnotesize,
  colback=black!3, colframe=black!55]
\emph{Per failure encountered}, record: \textbf{(1)} a stable bug
ID and its failure class (e.g.\ F1--F5 or a project-local code);
\textbf{(2)} discovery date and \emph{how} it was found (which
inspection or test surfaced it---not just ``code review'');
\textbf{(3)} affected cells (every model\,$\times$\,benchmark the
defect touched); \textbf{(4)} the symptom \emph{as it appeared in
the reported numbers} (what a reader would have seen and believed);
\textbf{(5)} provenance---\emph{inherited} from the initial
pipeline vs.\ \emph{introduced during repair}; \textbf{(6)} the fix
(the specific code change) \emph{and} a regression test that fails
on the pre-fix code; \textbf{(7)} the before/after delta on the
headline statistic for each affected cell; \textbf{(8)} residual
status: \textsc{fixed}, \textsc{disclosed-unfixed}, or
\textsc{out-of-scope}, with the reason. \emph{Per chronology as a
whole}, also state: total entries, how many were repair-introduced,
and which remain unfixed and why. An entry missing field (4) or (7)
is not auditable: the reader cannot tell what the wrong number was
or how much it moved.
\end{tcolorbox}

\paragraph{Why we report four buckets, not one verdict.} The four
non-confirmatory statuses in \S\ref{sec:results} separate different
consumer actions: \emph{ineligible} calls for scope revision,
\emph{scorer-unvalidated} for scorer validation, \emph{failed
G2--G4} for threshold or evidence revision, and \emph{exploratory}
for disclosed, non-confirmatory release. Collapsing these into a
single ``inconclusive'' verdict would hide the reasoning step that a
governance reviewer most needs.

\paragraph{Zero confirmatory cells is procedural, not an empirical
null.} The zero confirmatory verdict is not a claim about the five
benchmarks. Under \S\ref{par:gate}, even a cell whose CI is wholly
above 1 remains \emph{exploratory} while G5 or G6 is
\textsc{proposed}; the two TruthfulQA cells would upgrade only after
those gates are implemented. The contribution is the gate and
taxonomy, not a benchmark-validity verdict. \emph{No
benchmark-validity claim is issued for any cell.}

\section{Conclusion}
\label{sec:conclusion}

Perturbation-based benchmark-validity audits can be silently wrong
in at least five pipeline-level ways. We documented all five in a
single 10-cell case study, and caught one (F3c) only after
introducing it ourselves during repair. F1--F5 is an illustrative,
non-exhaustive starting list whose calibration across codebases is
open work.

The reusable artefact is the F1--F5 taxonomy, the
\S\ref{par:gate} status-assignment algorithm, the G1--G6 gate, and
the self-audit chronology of Box~1. This is a withholding and
disclosure protocol, not a scalar metric or benchmark leaderboard:
all 10 cells in our panel remain non-confirmatory, and even a future
\emph{confirmatory-selective} cell would certify contrast magnitude,
not signed direction (Appendix~\ref{app:limitations}).

Our ask is one sentence. \emph{Any audit producing benchmark-based
evidence for governance use should publish its self-audit
chronology---at minimum the eight fields of
Box~1---alongside its numbers.} Without it, a
clean number and a silently broken pipeline look the same.

\bibliographystyle{icml2026}
\bibliography{references}

\newpage
\appendix
\onecolumn
\section{Detailed Limitations}
\label{app:limitations}

We place limitations here so the reader can calibrate every number in
the main text against what this evidence cannot support.

\begin{enumerate}[leftmargin=*]
\item \textbf{Two models, five benchmarks, 200 items per cell, one
      seed, one prompt template.} Our 10-cell grid (Qwen-2.5-7B and
      Mistral-7B-v0.3 $\times$ five benchmarks) cannot support any
      claim that a class of 7--8B open-weight models exhibits any
      pattern.
\item \textbf{Post-prereg exploratory status.} Our preregistration
      (\S\ref{app:repro}) was committed before we discovered
      failures F3b and F3c; analyses after those discoveries are
      formally exploratory.
\item \textbf{Structurally-ineligible cells.} CrowS-Pairs is
      archetype-ambiguous under our PLL + demographic-swap
      configuration and additionally hits F1 scorer-path mismatch;
      BBQ is an invariance benchmark penalised by our diagnostic
      ratio metric (F5). Their numbers illustrate the two failure
      classes but are not benchmark-validity claims.
\item \textbf{Unvalidated scorers for ToxiGen and XSTest.} ToxiGen
      scoring uses targeted first-token log-probabilities over a
      fixed token set (after the F3c repair); XSTest refusal
      classification uses a ten-pattern regex over generations that
      we did not separately audit for false-negatives against human
      judgement. Neither scorer is a human-judged label; both cells
      are reported in the \emph{scorer-unvalidated} bucket.
\item \textbf{CSR is a contrastive-magnitude metric, not a
      signed-direction metric.} \csr{} uses unsigned family-deltas,
      so a paired-bootstrap CI wholly above 1 only certifies that
      $|\dattr|$ exceeds surface-family magnitudes. It does
      \emph{not} certify that attribute-flip responses are in the
      construct-consistent direction: a large $|\dattr|$ is
      consistent with any of construct-sensitive flipping,
      sycophancy flipping, lexical-salience noise, or an uncaught
      scorer artefact. Upgrading a \emph{confirmatory-selective}
      reading to \emph{construct-sensitive} requires signed-effect
      validation (the E3 human audit, deferred) and is future work.
\item \textbf{Canonical scorers are audit-hardened, not
      reference-faithful across the board.} TruthfulQA MC1 with
      deterministic per-item shuffle, ToxiGen targeted-token
      first-token logprobs, and XSTest's 10-pattern refusal regex
      are audit-robust substitutes rather than reference-faithful
      implementations; see Appendix~\ref{app:scoring} for the
      reference~/~audit~/~validation split. CrowS-Pairs PLL on
      paired sentences~\cite{salazar2020masked} and BBQ option
      log-probabilities split by \texttt{context\_condition} are
      reference-faithful.
\item \textbf{The six-point checklist is a proposed tool, not a
      standard.} It does not validate thresholds against external
      reference data, does not assign institutional ownership, and
      does not include anti-gaming provisions. Three further
      benchmark-specific limitations
      (shared-prefix MCQ tokens, \texttt{add\_instruction}
      contamination, weak surface edits on binary tasks) are listed
      in Appendix~\ref{app:bugs}.
\end{enumerate}

\section{Panel and Compute}
\label{app:panel}

Compute: a single NVIDIA RTX~4080 (16~GB) with HuggingFace
Transformers~4.57 in bfloat16; approximately 26 GPU-hours total
across the legacy and canonical pipelines. The panel originally
included Llama-3.1-8B; per author decision, Llama was dropped after
the legacy runs and before the canonical rerun. Only the two-model
canonical panel is the case study of this paper; legacy-Llama numbers
appear only in this appendix and the case-level audit trail
(Appendix~\ref{app:bugs}). Software versions:
Python~3.13, PyTorch~2.10, \texttt{transformers}~4.57.6; model
revisions pinned at load time
(\texttt{Qwen/Qwen2.5-7B-Instruct} and
\texttt{mistralai/Mistral-7B-Instruct-v0.3}; tokenizer revision
recorded in results JSON).

\section{Perturbation-Family Realisations}
\label{app:pertfam}

Three families per item; three perturbations per item per family.
Parse failures (legacy regex only) are symmetric: if either the
original or the perturbed item fails parsing, the item-level paired
difference for that family is dropped.
\emph{Format:} case changes, whitespace / punctuation noise,
label-format changes, choice reordering, and---for
text-classification benchmarks only---instruction prepend. We retain
\texttt{add\_instruction} in the format family despite its
task-boundary contamination on ToxiGen
(Appendix~\ref{app:bugs} marks this as unfixed-acknowledged).
\emph{Semantic:} paraphrase templates, synonym substitution, rule-based
sentence restructure.
\emph{Attribute:} benchmark-specific construct flips:
opposite-construct stem rewrite (TruthfulQA), demographic swap
(BBQ, CrowS-Pairs), toxicity-flip rewrite (ToxiGen),
safe$\leftrightarrow$unsafe pair lookup (XSTest).

\section{Scoring Pipelines}
\label{app:scoring}

The \emph{legacy} pipeline uses a uniform regex-extraction scorer over
free-form generation for all multiple-choice benchmarks and a
first-token logit comparison for ToxiGen and XSTest. The
\emph{canonical} pipeline implements each benchmark's intended
scoring protocol: option log-probabilities for TruthfulQA MC1, option
log-probabilities split by \texttt{context\_condition} for BBQ,
pseudo-log-likelihood over the paired sentences for CrowS-Pairs
(following \citet{salazar2020masked}), targeted-token first-token
log-probabilities for ToxiGen, and a ten-pattern regex refusal
classifier over generations for XSTest.

The legacy$\to$canonical fix list: read \texttt{label\_format} when
rendering MCQ prompts (F1 fix); replace \texttt{flip\_truthfulqa}
label inversion with opposite-construct stem rewrite; switch CrowS
scoring to PLL on paired sentences (F3a); switch TruthfulQA loading
from MC2 to MC1 to eliminate correct-first ordering (F3b); use
targeted-token first-token log-probabilities for ToxiGen to bypass
top-$k$ truncation (F3c); switch the bootstrap to an item-level
paired resampler (F4).

\paragraph{Reference vs.\ audit vs.\ validation evidence.}
``Canonical'' collapses three distinct things that should be kept
apart when reading downstream claims. Table~\ref{tab:scorer-fidelity}
splits them.

\begin{table}[h]
\centering
\footnotesize
\caption{Scorer fidelity split. \emph{Reference scorer} $=$ the
benchmark's published scoring protocol. \emph{Audit scorer used
here} $=$ the scorer our canonical pipeline actually runs.
\emph{Validation evidence} $=$ what we have to back the audit
scorer; \textsc{none} means the cell ships as
\emph{scorer-unvalidated} under \S\ref{par:gate}.}
\label{tab:scorer-fidelity}
\begin{tabular}{@{}p{0.14\linewidth}p{0.26\linewidth}p{0.26\linewidth}p{0.26\linewidth}@{}}
\toprule
Benchmark & Reference scorer & Audit scorer used here & Validation evidence \\
\midrule
TruthfulQA
& MC1 over options, accuracy
& MC1 over options with deterministic per-item shuffle (F3b fix)
& Reference-faithful up to the shuffle, which does not change the population metric. \\
\addlinespace
BBQ
& Option log-probabilities split by \texttt{context\_condition}
& Same
& Reference-faithful. \\
\addlinespace
CrowS-Pairs
& PLL on paired sentences~\cite{salazar2020masked}
& Same, reading the \texttt{choices} field
& Reference-faithful on the scored object; G1 instrumentation against the \emph{question}-edit path is not yet in place, which is why both cells ship as ineligible-F1. \\
\addlinespace
ToxiGen
& First-token logprob of \texttt{toxic}/\texttt{non-} tokens (post-F3c fix)
& Same, with targeted-token bypass of top-$k$
& Tokenizer-coverage probe on a 100-item subset only; \textsc{none} human-label. Ships as scorer-unvalidated. \\
\addlinespace
XSTest
& Refusal classification
& 10-pattern regex refusal classifier over generations
& \textsc{none}; no false-negative audit against human labels. Ships as scorer-unvalidated. \\
\bottomrule
\end{tabular}
\end{table}

\section{Threshold Honesty}
\label{app:thresholds}

The $0.02$ threshold in G3 and the $0.01$ floor of $\varepsilon$ are
\emph{engineering choices}, not statistical thresholds. The intent is
to avoid ratio explosion when both surface-family deltas are near
zero; the specific numeric choice is tied to our 200-item subset size
but we do not claim it is optimal or that $0.02$ equals ``two standard
errors'' of anything (for a proportion at $p=0.5$ with $n=200$,
$2\,\mathrm{SE}\approx 0.07$, so $0.02$ is substantially more
permissive than a strict statistical-significance threshold). The
$20\%$ silent-no-op cap in G1 is likewise pragmatic, motivated by our
observed 0--83\% range (Appendix~\ref{app:autodiff}). A fuller
treatment would derive each threshold from a target sensitivity level
in the downstream evidence consumer. Satisfying G1--G6 does not
establish that any benchmark is construct-valid; it substantially
reduces the risk that non-confirmation is driven by known preventable
pipeline failures. A production due-diligence instrument would
additionally require institutional ownership, externally-validated
thresholds, anti-gaming provisions, and legal-compatibility review.

\paragraph{Threshold sensitivity of the 10-cell breakdown.}
We resweep the three threshold knobs and re-derive the five-bucket
counts of \S\ref{sec:results} under each setting; the qualitative
breakdown is stable over the grid we considered.
\textbf{G3} in $\{0.01, 0.02, 0.05\}$: the only boundary cell is
Mistral\,$\times$\,XSTest with
$\max(|\dfmt|,|\dsem|)\!=\!0.012$. At $0.01$ that cell passes G3
but still fails G2 ($0.565$ vs.\ $0.500$, under $2\,\mathrm{SE}$),
so it remains \emph{failed G2--G4}; at $0.05$ it fails both G2 and
G3 and remains \emph{failed G2--G4}. Every other in-panel cell with
non-degenerate denominator has $\max(|\dfmt|,|\dsem|)\!\geq\!0.050$,
and the F1-degenerate CrowS cells are pre-filtered.
\textbf{$\varepsilon$} in $\{0.001, 0.01, 0.1\}$: $\varepsilon$ only
floors the CSR denominator; it shifts CSR point estimates on
denominator-dominated cells but does not change any G2 or G3
outcome, so bucket assignment is unchanged. \textbf{G1 silent-no-op
cap} in $\{5, 10, 20\}\%$: the observed canonical-scorer no-op rates
are $\{0.0, 0.0, 100, 100, 83\}\%$ for
XSTest/ToxiGen/TruthfulQA/BBQ/CrowS; any cap $\leq 83\%$ leaves
CrowS F1-ineligible and TruthfulQA/BBQ F1-passing (their no-op
rates are $0\%$ on the canonical path after the F1b repair). The
$3/3/2/2/0$ breakdown therefore holds over this grid. Thresholds
outside the grid (e.g., G3 $=0.1$, which would flag denominators
up to half the $p\!=\!0.5,n\!=\!200$ 2-SE of $0.07$) would
reclassify TruthfulQA's exploratory cells; we report the grid above
as the regime where the thresholds are defensible as conservative
engineering floors.

\section{Per-Benchmark Eligibility Notes}
\label{app:eligibility}

The main-body eligibility breakdown (\S\ref{sec:results},
Table~\ref{tab:eligibility}, Figure~\ref{fig:forest-main}) summarises
per-cell status under the \S\ref{par:gate} precedence. Additional
per-benchmark notes:

\textbf{TruthfulQA.} Both cells score above the $1/4$ uniform
baseline (G2 passes: Qwen $0.485$, Mistral $0.520$) with
$\max(|\dfmt|,|\dsem|)\!\geq\!0.0767$ (G3 passes) and paired
bootstrap CIs wholly above 1 (Qwen CSR $13.04$, CI $[10.53, 17.65]$;
Mistral CSR $8.33$, CI $[6.67, 10.34]$). Both cells therefore land
in the \emph{exploratory} bucket pending G5 and G6 enforcement.
\textbf{BBQ.} Mistral\,$\times$\,BBQ passes all implemented gates
but is \emph{ineligible (F5)} because the attribute family swaps
demographic tokens while the answer key stays fixed, so $\dattr$
is not a clean construct flip. Qwen\,$\times$\,BBQ fails G2
(unperturbed score $0.340$ vs.\ $1/3$ baseline is $0.7\,\mathrm{pp}$
above, well under the $2\,\mathrm{SE}\approx 6.7\,\mathrm{pp}$
margin) and additionally inherits the F5 concern; it falls
in the \emph{failed G2--G4} bucket under first-match precedence.
\textbf{CrowS-Pairs.} Both cells are \emph{not\_applicable} (mapped
to \emph{ineligible} in paper): the canonical PLL scorer reads
\texttt{choices}, while the format / semantic perturbations edit
\texttt{question}; the perturbation does not reach the scored
object (F1), and archetype is ambiguous under PLL + question-edit.
$|\dsem|\!=\!0$ on both cells is consistent with F1.
\textbf{ToxiGen.} Both cells pass G1--G4 but the scorer (first-token
logprob over \texttt{toxic}/\texttt{non-} tokens) is validated only
on a 100-item tokenizer-coverage probe (Appendix~\ref{app:panel};
mean coverage $0.853$ and $0.857$, $p_5 = 0.64$ and $0.74$).
Absent a held-out human-judged validation, both ToxiGen cells are
labelled \emph{scorer-unvalidated}.
\textbf{XSTest.} The 10-pattern refusal regex has not been separately
audited for false-negative rate. Qwen\,$\times$\,XSTest passes all
implemented gates (\emph{scorer-unvalidated});
Mistral\,$\times$\,XSTest fails G2 above-baseline \emph{and} is
denominator-dominated at $\max(|\dfmt|,|\dsem|)\!=\!0.012<0.02$
(G3 fails), falling in \emph{failed G2--G4}.

The canonical 200-item ToxiGen slice is class-balanced, so the
uniform and majority baselines coincide at $0.5$. Gate decisions use
full precision throughout; displayed values may round to the
threshold.

\paragraph{Seed stability and placebo specificity (PREREG E5, E6).}
Of the 9 cells with non-degenerate CSR (Qwen-2.5-7B $\times$ BBQ
fails G2 above-baseline, so its seed CV is undefined), per-cell
CSR coefficient of variation across seeds $\{42, 123, 2024\}$ is
$\leq 0.27$ on all 9 (smallest $0.021$ on Qwen\,$\times$\,TruthfulQA;
largest $0.266$ on Mistral\,$\times$\,XSTest). Seed instability
does not drive any verdict. The placebo specificity ratio
$\mathrm{CSR}_{\mathrm{real}} / \mathrm{CSR}_{\mathrm{placebo}}$
is $\geq 4$ on 4 of the 8 cells with $\mathrm{CSR}_{\mathrm{real}} > 0$
(Qwen\,$\times$\,XSTest $5.00$, Qwen\,$\times$\,ToxiGen $4.47$,
Mistral\,$\times$\,XSTest $74.25$, Mistral\,$\times$\,ToxiGen
$5.61$; median across all 8 cells is $3.98$). Excluding the two
BBQ cells where $\mathrm{CSR}_{\mathrm{real}}\!\approx\!0$ so the
ratio is denominator-noise dominated (raw ratios $0.017$ and
$0.165$), the ratio is $\geq 4$ on 4 of 6 cells, median $4.74$.
This is positive but non-universal support for contrast
specificity. Source data:
\texttt{results/canonical/seed\_variance.csv} and
\texttt{results/canonical/csr\_placebo.csv}.

\section{Verdict-Mapping Rule}
\label{app:verdict-mapping}

The raw \texttt{cell\_validity.csv} \texttt{verdict} column encodes,
per cell, the first condition that fires under the status-assignment
algorithm of \S\ref{par:gate}. The raw-verdict tokens and the
algorithm steps that emit them are:
\texttt{not\_applicable} (G1 uninstrumented, step 1),
\texttt{inconclusive} (at least one of G2/G3/G4 fails, step 2),
\texttt{uninterpretable\_scorer} (gates pass, scorer not validated,
step 3),
\texttt{attribute\_family\_mismatch} (gates pass and scorer
validated, but archetype-mismatched under CSR-as-defined, step 4),
and \texttt{selective} / \texttt{non\_selective} (all of the above
pass; CI wholly above / below 1; step 5 or 6). Per
\S\ref{par:gate}, the raw-verdict tokens are emitted in the order
they are checked; a reader with the raw CSV can reproduce every
count in the paper by applying the table below mechanically.

\begin{table}[h]
\centering
\footnotesize
\caption{Raw \texttt{cell\_validity.csv} \texttt{verdict} $\to$
paper bucket. The raw-verdict emission order matches
\S\ref{par:gate} exactly, so no first-match contention occurs at
the mapping step. A cell with \texttt{verdict=selective} is
downgraded to \emph{exploratory} in this submission because G5 and
G6 remain \textsc{proposed}; it would upgrade to
\emph{confirmatory-selective} once both gates are
\textsc{implemented}.}
\label{tab:verdict-mapping}
\begin{tabular}{lll}
\toprule
Raw \texttt{verdict} & Paper bucket & \S\ref{par:gate} step \\
\midrule
\texttt{not\_applicable}                       & ineligible (F1)              & 1 \\
\texttt{inconclusive}                          & failed G2--G4                 & 2 \\
\texttt{uninterpretable\_scorer}               & scorer-unvalidated           & 3 \\
\texttt{attribute\_family\_mismatch}           & ineligible (F5)              & 4 \\
\texttt{selective} (G5 / G6 \textsc{proposed}) & exploratory                  & 5 \\
\texttt{non\_selective} (G5 / G6 \textsc{proposed}) & exploratory             & 5 \\
\texttt{selective} (G5 / G6 \textsc{implemented}) & confirmatory-selective    & 6 \\
\texttt{non\_selective} (G5 / G6 \textsc{implemented}) & confirmatory-non-selective & 6 \\
\bottomrule
\end{tabular}
\end{table}

\paragraph{Worked example: Mistral\,$\times$\,XSTest.}
Raw verdict \texttt{inconclusive}; algorithm trace:
step 1 fires G1? No---XSTest's 10-pattern refusal regex reads the
generated text and our format perturbations do reach that path
(G1 is instrumented, though the scorer itself is unvalidated; that
is step 3's concern, not step 1's).
Step 2: G2 fails (unperturbed $0.565 <$ $0.500 + 2\,\mathrm{SE}$)
\emph{and} G3 fails
($\max(|\dfmt|,|\dsem|)\!=\!0.012 < 0.02$). Step 2 fires;
raw verdict \texttt{inconclusive}; paper bucket
\emph{failed G2--G4}. Step 3 (scorer-unvalidated) would have
fired had G2--G4 passed, but it did not. The scorer-unvalidated
flag is still surfaced in the appendix eligibility notes as a
secondary concern.
\paragraph{Worked example: Qwen\,$\times$\,BBQ.} Raw verdict
\texttt{inconclusive}; step 1 does not fire (BBQ option-logprob
scorer is fully instrumented); step 2 fires because G2 fails
($0.340 <$ $0.333 + 2\,\mathrm{SE}$); bucket \emph{failed
G2--G4}. The F5 archetype-mismatch concern is benchmark-level and
would have fired at step 4 had G2--G4 passed
(as on Mistral\,$\times$\,BBQ).

\paragraph{Cell-level \texttt{gate\_*} columns vs.\ paper G1--G6.}
The raw \texttt{cell\_validity.csv} gate columns
(\texttt{gate\_parseability}, \texttt{gate\_above\_baseline},
\texttt{gate\_denom\_not\_eps}, \texttt{gate\_ci\_estimable},
\texttt{gate\_n\_items\_ok}, \texttt{gate\_seed\_cv\_lt\_03}) are
the empirical per-cell instantiation of the methodology-level
G1--G6 in \S\ref{sec:method}, not a separate gate set.
\texttt{gate\_parseability} realises the regex-parseability half
of G1; the silent-no-op half of G1 is audited offline against the
canonical scorer's prompt builder
(Table~\ref{tab:autodiff}), which is why G1 ships as \textsc{partial}
in Table~\ref{tab:checklist}.
\texttt{gate\_above\_baseline} realises G2;
\texttt{gate\_denom\_not\_eps} realises G3;
\texttt{gate\_ci\_estimable} + \texttt{gate\_n\_items\_ok} +
\texttt{gate\_seed\_cv\_lt\_03} jointly realise G4 (paired-bootstrap
feasibility, sample-size guard, seed-stability guard). The CSV
carries no direct column for G5 (archetype disclosure) or G6
(repair-regression); their \textsc{proposed} status is what
downgrades a raw-verdict \texttt{selective} to paper-bucket
\emph{exploratory} under \S\ref{par:gate}.

\section{Case-Study Table: 10 Cells, Canonical Pipeline}
\label{app:casetable}

Table~\ref{tab:cell-validity} is the per-cell evidence behind the
counts in \S\ref{sec:discussion} / Table~\ref{tab:eligibility}. All
values are taken directly from
\texttt{results/canonical/csr\_canonical.csv} (per-family deltas,
CSR point estimate) and
\texttt{results/canonical/cell\_validity.csv} (paired-bootstrap CI,
raw verdict). Gate decisions use full precision throughout.

\begin{table*}[h]
\centering
\small
\caption{Canonical-pipeline case-study numbers, Qwen-2.5-7B and
  Mistral-7B-v0.3. Raw verdicts are the literal
  \texttt{cell\_validity.csv} values; the paper-bucket mapping is
  Appendix~\ref{app:verdict-mapping}. Baselines: TruthfulQA MC1
  $1/4$, BBQ $1/3$, ToxiGen $0.5$, XSTest $1/2$.}
\label{tab:cell-validity}
\begin{tabular}{llrrrrrrl}
\toprule
Model & Benchmark & $s_{\text{orig}}$ & $\dfmt$ & $\dsem$ & $\dattr$ &
\csr{} & 95\% CI & Raw verdict \\
\midrule
Qwen-2.5-7B  & CrowS-Pairs  & 0.655 & 0.095 & 0.000 & 0.135 &  1.42 & [1.01,  2.06] & \texttt{not\_applicable} \\
Qwen-2.5-7B  & BBQ          & 0.340 & 0.050 & 0.000 & 0.000 &  0.00 & [0.00,  0.00] & \texttt{inconclusive} \\
Qwen-2.5-7B  & XSTest       & 0.620 & 0.140 & 0.140 & 0.817 &  5.83 & [4.31,  7.23] & \texttt{uninterpretable\_scorer} \\
Qwen-2.5-7B  & TruthfulQA   & 0.485 & 0.077 & 0.000 & 1.000 & 13.04 & [10.53, 17.65] & \texttt{selective} \\
Qwen-2.5-7B  & ToxiGen      & 0.785 & 0.058 & 0.087 & 0.708 &  8.17 & [6.04, 11.53] & \texttt{uninterpretable\_scorer} \\
Mistral-7B   & CrowS-Pairs  & 0.670 & 0.095 & 0.000 & 0.133 &  1.40 & [0.99,  1.98] & \texttt{not\_applicable} \\
Mistral-7B   & BBQ          & 0.425 & 0.113 & 0.088 & 0.038 &  0.34 & [0.20,  0.51] & \texttt{attribute\_family\_mismatch} \\
Mistral-7B   & XSTest       & 0.565 & 0.012 & 0.008 & 0.990 & 84.86 & [44.92, 99.83] & \texttt{inconclusive} \\
Mistral-7B   & TruthfulQA   & 0.520 & 0.120 & 0.105 & 1.000 &  8.33 & [6.67, 10.34] & \texttt{selective} \\
Mistral-7B   & ToxiGen      & 0.765 & 0.062 & 0.053 & 0.692 & 11.22 & [7.82, 15.97] & \texttt{uninterpretable\_scorer} \\
\bottomrule
\end{tabular}
\end{table*}

\begin{figure}[h]
  \centering
  \includegraphics[width=0.85\linewidth]{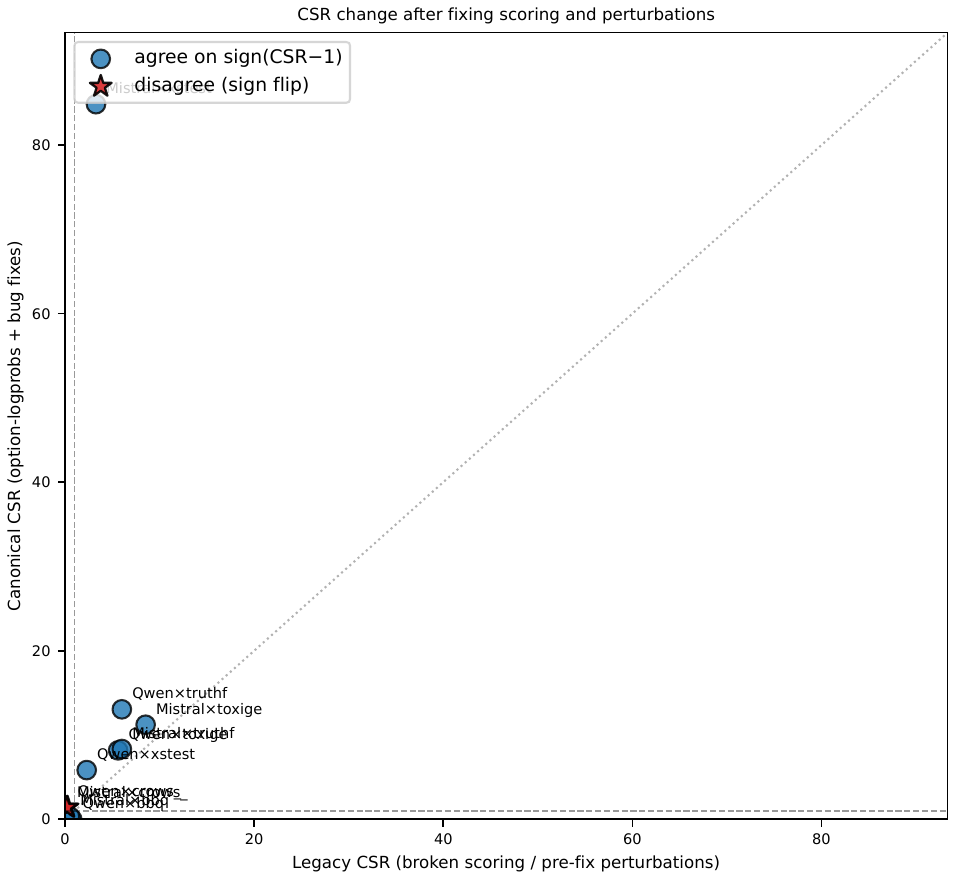}
  \caption{Legacy-pipeline CSR (x-axis) versus canonical-pipeline
    CSR (y-axis) per cell, on \emph{symlog} axes (linear in
    $[-1,1]$, log outside) so both the near-zero BBQ cells and the
    Mistral\,$\times$\,XSTest outlier at CSR $\approx 85$ are
    readable on one panel. Points on the dotted identity line are
    unchanged by repair; points far from it reflect the joint
    effect of F1--F3 fixes. Seven of ten cells change verdict
    (selective~$\leftrightarrow$~non-selective or
    denominator-dominated) under the repaired pipeline. Cell
    codes: Q $=$ Qwen-2.5-7B, M $=$ Mistral-7B-v0.3; TQ $=$
    TruthfulQA, BB $=$ BBQ, TX $=$ ToxiGen, CP $=$ CrowS-Pairs,
    XS $=$ XSTest.}
  \label{fig:legacy-vs-canonical}
\end{figure}

\begin{figure}[h]
  \centering
  \includegraphics[width=0.85\linewidth]{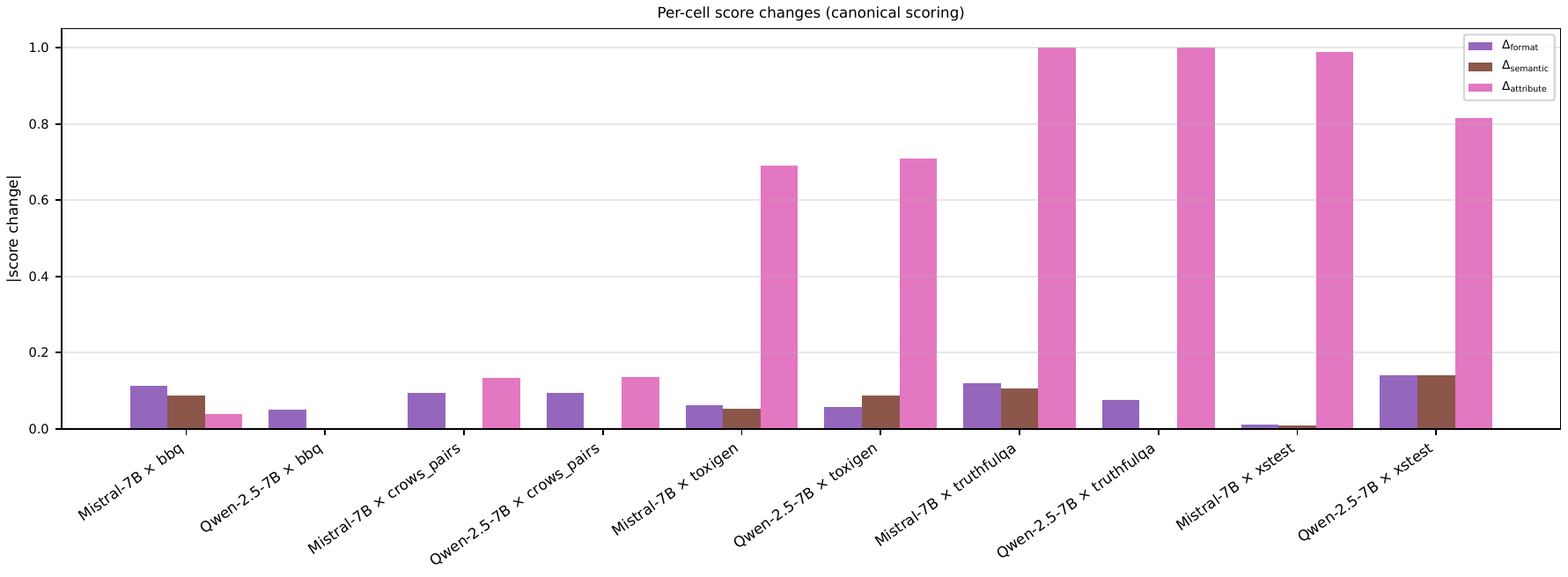}
  \caption{Per-cell $|\dfmt|$, $|\dsem|$, $|\dattr|$ under the
    canonical pipeline. Cells where $|\dfmt|$ or $|\dsem| < 0.02$
    drive the denominator below the $\varepsilon$ floor and trigger
    G3. $|\dsem| = 0$ on CrowS-Pairs reflects F1-type silent no-ops on
    the question field, which the PLL scorer does not read.}
  \label{fig:deltas}
\end{figure}

\clearpage

\section{Auto-Diff Audit Numbers}
\label{app:autodiff}

Silent-no-op rates of format perturbations per benchmark, measured
(i) against the legacy renderer
(\texttt{evaluate.format\_mcq\_prompt}) and (ii) against the canonical
scorer's prompt builder
(\texttt{scoring\_canonical.\_format\_mcq\_prompt}). The gap between
the two columns is the core F1 evidence.

\begin{table}[h]
\centering
\small
\caption{Silent-no-op rate (\%) for the format family. Column L
$=$ against the legacy renderer; column C $=$ against the canonical
scorer's prompt builder. The F1 regression shows in the C column for
MCQ benchmarks (100\% no-op).}
\label{tab:autodiff}
\begin{tabular}{lrr}
\toprule
Benchmark   & L (legacy) & C (canonical MCQ) \\
\midrule
TruthfulQA  &  0.2  & 100   \\
BBQ         &  1.8  & 100   \\
CrowS-Pairs & 11.0  &  83\,$^\dagger$ \\
ToxiGen     &  0.0  & --    \\
XSTest      &  0.0  & --    \\
\bottomrule
\end{tabular}
\par\smallskip
\footnotesize
$^\dagger$ CrowS under PLL: 5 of 6 perturbation methods do not reach
the scorer because the scorer reads \texttt{choices} and the
perturbations edit \texttt{question}. The 83\% figure is the
effective scorer-consumed no-op rate.
\end{table}

\section{Case-Level Audit Trail}
\label{app:bugs}

This appendix is the worked instance of the minimal chronology
template (Box~1): each entry below carries a
bug ID and failure class (field 1), provenance (field 5: F3c is
flagged \emph{self-introduced}), the fix and its regression test
(field 6: tests shipped for F1, F3c, F4), and residual status
(field 8: \textsc{fixed} / \textsc{unfixed, acknowledged}). For
space, the per-cell discovery date, affected-cell list, and
before/after deltas (fields 2--4, 7) are condensed here and given
in full in the release bundle's running audit log
(Appendix~\ref{app:repro}); the F4 entry quotes its before/after
delta inline as an example of field 7. All items listed as
\textsc{fixed} are repaired in the canonical pipeline that
produced every number in this paper; the trail exists so a reader
can audit our audit, not because the canonical pipeline ships with
unresolved bugs. Three items remain \textsc{unfixed, acknowledged}
and are disclosed as active limitations.

Entries are chronological, matched to the failure-mode cases in
\S\ref{sec:cases}.

\begin{enumerate}
\item \textbf{F1a (fixed)}: \texttt{format\_change\_labels} wrote
  \texttt{item[`label\_style']}; prompt builder did not read it.
  Fix: parametrise \texttt{format\_mcq\_prompt(label\_style)}.
\item \textbf{F1b (fixed)}: Fix landed only in \texttt{evaluate.py};
  \texttt{scoring\_canonical.py::\_format\_mcq\_prompt} remained
  hard-coded. Fix: parametrise the canonical builder.
\item \textbf{F1c (fixed)}: Rule-based
  \texttt{semantic\_sentence\_restructure} had no-op rates of 57--93\%
  due to trigger-string mismatch on Q\&A benchmarks. Fix: add
  \texttt{paraphrase\_template}, \texttt{synonym\_substitution}.
\item \textbf{F2a (partially fixed)}: Qwen $\times$ TruthfulQA legacy
  parseability 0.44. Fix: canonical option-logprob scoring sets
  parseability to 1.0; legacy numbers are withdrawn.
\item \textbf{F3a (disclosed, not repaired in data)}: CrowS-Pairs
  legacy accuracy inverts fairness interpretation. Canonical pipeline
  switches to PLL; CrowS legacy numbers are withdrawn.
\item \textbf{F3b (fixed)}: TruthfulQA MC2 correct-first ordering
  artefact. Fix: MC1 + per-item shuffle.
\item \textbf{F3c (fixed, self-introduced)}: Top-$k$ truncation caused
  majority-class predictions on ToxiGen. Fix: targeted-token
  first-token log-probabilities.
\item \textbf{F4 (fixed)}: Broken bootstrap pairing. Fix: item-level
  paired bootstrap. Out-of-panel legacy-Llama at $\csr \approx 9.5$
  showed CI width 26.4 under unpaired resampling.
\item \textbf{F5 (disclosed)}: BBQ / CrowS archetype mismatch under
  CSR. Not fixed; G3 gate flags denominator-dominated cells.
\item \textbf{Shared-prefix collapse on MCQ letter tokens}
  (\textsc{unfixed, acknowledged}): first token of ``\texttt{(A)}'',
  ``\texttt{(B)}'' can be identical.
\item \textbf{TEXTCLF \texttt{add\_instruction} content contamination
  on ToxiGen} (\textsc{unfixed, acknowledged}).
\end{enumerate}

\paragraph{Containment of unfixed issues.}
The remaining unfixed items do not alter any \emph{confirmatory}
verdict in this paper because no confirmatory verdict is issued: all
affected cells remain non-confirmatory under the hierarchical
precedence of \S\ref{par:gate}. The unfixed items therefore further
motivate---rather than undercut---the paper's withholding rule.

\section{Preregistration and Release}
\label{app:repro}

\textbf{Code and data.} A release bundle will accompany the
de-anonymised version of the paper, containing the canonical
pipeline (all F1--F4 fixes landed), per-item predictions for all
10 cells, the preregistration \texttt{PREREG.md} (commit
\texttt{e6e686b8c0b02dd78d1495eb9c4a94882cce5cce}, amended
2026-04-22), running audit logs, and regression tests for F1, F3c,
and F4. The bundle is not part of this submission and the URL is
withheld for blind review. Hardware and software versions are in
Appendix~\ref{app:panel}.

\textbf{Deviations from preregistration.} The 2026-04-19 prereg
was amended on 2026-04-22 (PREREG \S10). The substantive
deviations:
\begin{enumerate}[leftmargin=1.4em,topsep=2pt,itemsep=1pt,label=(\alph*)]
\item Llama-3.1-8B was dropped between prereg and canonical rerun,
  reducing the panel to two models.
\item F3b and F3c were discovered after the prereg SHA; all
  analyses after their discovery are formally exploratory.
\item The E3 human audit was de-scoped; the metric was renamed
  ``Construct Sensitivity Ratio'' $\to$ ``Contrast Selectivity
  Ratio'' under the pre-committed PREREG \S7 fallback.
  Construct-validity language is retracted from any claim that
  would have required E3 validation; artifacts
  \texttt{results/audit/}, \texttt{ETHICS.md}, and
  \texttt{annotator\_consent.md} are removed from the PREREG \S9
  manifest.
\item CrowS-Pairs is marked \emph{not\_applicable} at the cell
  level (the PLL scorer reads \texttt{choices} while our
  perturbations edit \texttt{question}); no
  selective/non\_selective verdict is issued for CrowS. See
  Appendix~\ref{app:eligibility}.
\item The legacy unpaired bootstrap
  (\texttt{src/ablations.py::bootstrap\_csr}) was removed;
  claim-bearing CIs come from
  \texttt{src/analyze\_canonical.py::paired\_bootstrap\_csr}
  (item-level paired, PREREG \S6.1). See \S\ref{sec:cases} (F4).
\item Template robustness (PREREG \S6.3) was scoped to TruthfulQA
  and ToxiGen and reported as secondary, not as a gate for the
  other three benchmarks whose scoring paths have no
  prompt-template axis.
\item The \texttt{cell\_validity.csv} gate set was tightened from
  3 columns (2026-04-21) to the full 6-column G1--G4 realisation;
  mapping to paper G1--G6 is in
  Appendix~\ref{app:verdict-mapping}.
\end{enumerate}

\textbf{Seed stability and placebo specificity (E5, E6).}
Reported in Appendix~\ref{app:eligibility}.

\end{document}